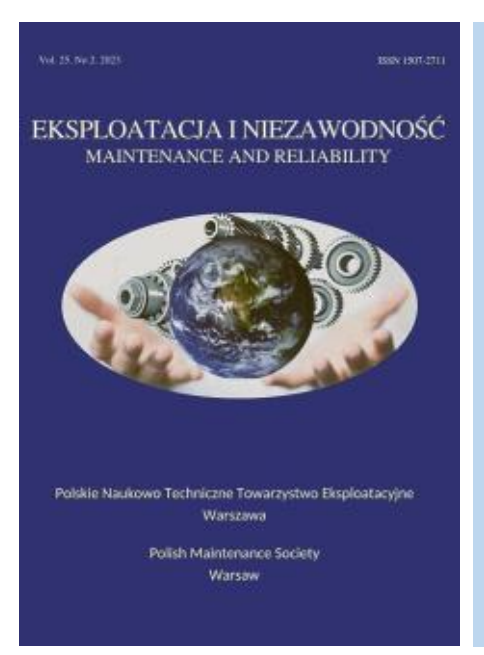

# Charging phase health indicators for battery state-of-health estimation: a systematic comparison of CC, CV, and combined approaches under cross-battery validation

**Huy Hoang Le[a], Kim-Anh Nguyen[b,*]**

[a] R&D Department, PowerMore Ltd., Viet Nam
[b] Faculty of Electrical Engineering, University of Science and Technology, The University of Danang, Viet Nam

## Highlights

- Comparison of CC-only, CV-only, and combined indicators under LOBO validation.
- Combined CC+CV yields the best SOH accuracy, showing complementary aging information.
- Robust evaluation via Leave-One-Battery-Out validation and SHAP analysis.
- 119% performance gap between cross-validation and realistic cross-battery evaluation.

## Abstract

Accurate State-of-Health estimation is essential for safe battery operation and cost-effective maintenance. Although numerous health indicators have been derived from constant-current (CC) and constant-voltage (CV) charging phases, their effectiveness under realistic cross-battery validation remains insufficiently studied. This work addresses this gap through a systematic comparison of CC-only, CV-only, and combined indicator sets using rigorous Leave-One-Battery-Out (LOBO) validation on the NASA battery aging dataset. Four CV-phase indicators and CC phase duration are evaluated individually and in combination. Results show that the combined CC+CV approach achieves the best performance ($R^2$ = 0.874), confirming that CC and CV phases capture complementary degradation information. Moreover, a 119% performance gap is observed between standard 5-fold cross-validation and LOBO validation, indicating that conventional evaluation overestimates practical accuracy. Based on these findings, practical guidelines are provided for indicator selection under data and computational constraints.

## 1. Introduction

Lithium-ion batteries have become the dominant energy storage technology in electric vehicles, renewable energy systems, and portable electronics due to their high energy density, long cycle life, and declining costs [1]. However, battery performance inevitably degrades over time due to complex electrochemical aging mechanisms, necessitating accurate State-of-Health (SOH) estimation for safe operation and optimal maintenance scheduling [2].

SOH estimation methods can be broadly categorized into model-based approaches, which rely on electrochemical or equivalent circuit models, and data-driven approaches, which learn degradation patterns directly from operational data [3]. While model-based methods provide physical interpretability, they require accurate parameterization and significant computational resources. Data-driven methods, particularly deep learning approaches, have demonstrated impressive accuracy but often demand substantial training data and computational power that may not be available in resource-constrained embedded systems [4].

From a maintenance engineering perspective, the primary purpose of SOH estimation is not necessarily to achieve the highest possible accuracy, but rather to provide reliable indicators for condition-based maintenance decisions [5]. For scheduling battery replacement or adjusting operating parameters, an estimation accuracy of several percent is often sufficient, provided the method is robust and practically implementable.

(*) Corresponding author.
E-mail addresses: H.H. Le (ORCID: 0009-0009-2138-5468) lhhoang@powermore.vn, K-A. Nguyen (ORCID: 0000-0003-3408-847X) nkanh@dut.udn.vn

The constant-current constant-voltage (CC-CV) charging protocol is widely adopted in battery management systems (BMS) due to its simplicity and effectiveness in balancing charging speed with battery longevity [6]. During CC-CV charging, the battery is first charged at constant current until reaching a voltage threshold, then held at constant voltage while the current gradually decreases until reaching a cutoff value. The CV phase behavior is directly influenced by battery internal resistance and diffusion characteristics, both of which change with aging [7].

Previous studies have primarily focused on features extracted from the constant-current (CC) phase, such as incremental capacity analysis (ICA) and differential voltage analysis (DVA) [8,9]. However, these methods require numerical differentiation of noisy measurement data, potentially amplifying measurement errors. The CV phase, in contrast, exhibits smooth current decay that can be characterized without differentiation operations.

This study addresses the lack of systematic comparison between different charging phase indicators under realistic cross-battery validation. The main contributions are:

1. Systematic comparison of CC-only, CV-only, and combined indicator sets for SOH estimation, revealing that the combined approach captures complementary degradation information from both charging phases.
2. Quantification of the substantial performance gap (119% RMSE increase) between standard cross-validation and Leave-One-Battery-Out (LOBO) validation, demonstrating that conventional evaluation methods significantly overestimate deployment accuracy.
3. SHAP-based interpretability analysis identifying the CV-to-CC time ratio as the most influential indicator, providing practical guidance for indicator selection in battery management systems.

The remainder of this paper is organized as follows. Section 2 reviews related work on battery SOH estimation. Section 3 describes the proposed methodology including indicator extraction and model development. Section 4 presents experimental results and analysis. Section 5 discusses the findings and limitations. Section 6 concludes the paper.

## 2. Related work

### 2.1. Data-driven SOH estimation

Data-driven approaches for battery SOH estimation have gained significant attention due to their ability to capture complex degradation patterns without requiring detailed electrochemical knowledge. Machine learning methods including support vector regression (SVR), Gaussian process regression (GPR), and ensemble methods have been successfully applied to SOH prediction [10,11].

Deep learning architectures, particularly long short-term memory (LSTM) networks and convolutional neural networks (CNN), have achieved state-of-the-art accuracy with reported RMSE values below 1% on benchmark datasets [12,13]. Recent advances include hybrid CNN-GRU architectures with error correction achieving MAE below 0.5% [14] and deep belief network-based methods for RUL prediction [15]. To address data scarcity challenges, generative approaches such as TimeGAN-based data augmentation have been explored for improving SOH prediction with limited samples [16]. However, these methods typically require large training datasets, significant computational resources, and may suffer from poor generalization to batteries with different chemistries or operating conditions [17]. For maintenance applications where computational resources are limited and real-time processing is required, lightweight machine learning models such as random forests and gradient boosting machines offer a practical compromise between accuracy and complexity [18].

### 2.2. Charging curve-based health indicators

The charging voltage and current profiles contain rich information about battery health status. Incremental capacity analysis (ICA), which examines the derivative dQ/dV, has been extensively studied for identifying aging mechanisms and estimating SOH [8]. However, ICA requires high-resolution voltage measurements and careful noise filtering to obtain reliable results. Several studies have explored time-domain features from charging curves. The charging time to reach specific voltage thresholds has been shown to correlate with capacity fade [19]. The area under the charging curve and statistical features of voltage profiles have also been investigated as health indicators [20]. Su et al. [21] proposed combining multiple health indicators derived from current, voltage, and temperature data with neural network-based capacity estimation. Burzynski [22] developed a useful energy prediction model for lithium-ion cells using machine learning techniques that account for various duty cycles and degradation patterns.

### 2.3. CV phase analysis

The CV charging phase has received relatively less attention compared to the CC phase, despite its potential advantages. During CV charging, the current decay profile reflects the battery's ability to accept charge, which is influenced by internal resistance and solid-state diffusion limitations [7].

Yang et al. [23] demonstrated that the CV phase time constant, obtained by fitting an exponential decay model to the current profile, shows linear correlation with capacity fade for $LiFePO_4$ batteries. Chen et al. [24] proposed using the ratio of CV charging time to CC charging time as a health indicator, noting that this ratio increases as the battery ages due to increased internal resistance. More recently, Ko et al. [25] achieved remarkable accuracy (MAE < 0.2%) using differential current analysis during the CV phase. However, their method still requires differentiation operations and may be sensitive to measurement noise in practical applications.

The present study builds upon these findings by systematically evaluating multiple CV phase indicators and assessing their practical applicability for maintenance-oriented SOH estimation.

## 3. Methodology

### 3.1. Dataset description

This study utilizes the publicly available NASA battery aging dataset [26], which contains cycling data from 18650-format lithium-ion cells with LiCoO2 cathode chemistry. Four batteries (B0005, B0006, B0007, B0018) subjected to charge-discharge cycling at room temperature are selected for analysis. Figure 1 illustrates the capacity degradation trajectories of the four lithium-ion batteries over cycling, with the commonly used 80% SOH end-of-life threshold indicated for reference.

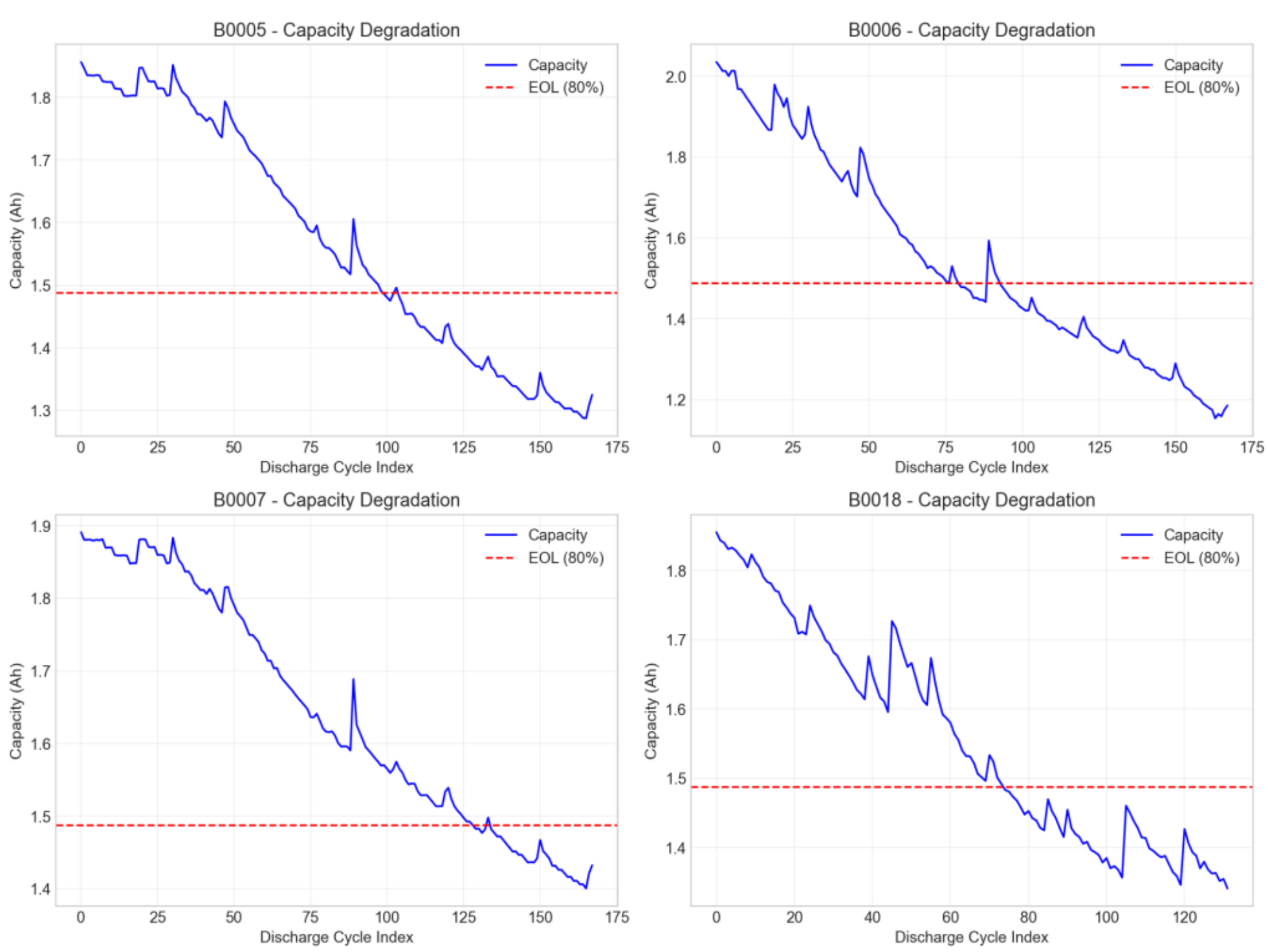

Figure 1. Capacity degradation curves for the four batteries (B0005, B0006, B0007, and B0018).

The charging protocol consists of CC charging at 1.5 A until the voltage reaches 4.2 V, followed by CV charging at 4.2 V until the current drops to 20 mA. Discharge is performed at 2 A constant current until the voltage reaches 2.7 V, with capacity measured by coulomb counting. The nominal capacity is determined as 1.86 Ah based on the maximum observed discharge capacity. SOH is calculated as the ratio of measured discharge capacity to nominal capacity, expressed as a percentage. Table 1 summarizes the dataset characteristics.

Regarding the nominal capacity definition: using the observed maximum (1.86 Ah), 38 cycles (6.1%) exhibit SOH values above 100%, concentrated in B0006 (18 cycles, max 104.6%) and B0007 (20 cycles, max 101.7%). These reflect natural cell-to-cell variation rather than measurement error. Comparison with the manufacturer-rated capacity (2.0 Ah) yields LOBO RMSE = 4.363% vs 4.692%, but identical $R^2$ = 0.7956, confirming that the nominal capacity definition does not affect relative predictive ability — the RMSE difference reflects only SOH scale compression.

Table 1. Summary of NASA battery aging dataset used in this study.

| Battery | Charge cycles | Discharge cycles | SOH range (%) |
|---|---|---|---|
| B0005 | 170 | 168 | 69.2 – 99.5 |
| B0006 | 158 | 156 | 69.8 – 99.6 |
| B0007 | 170 | 168 | 70.1 – 99.4 |
| B0018 | 132 | 131 | 69.0 – 104.6 |
| Total | 630 | 623 | 62.0 – 104.6 |

### 3.2. CV phase detection

The CV phase is identified within each charging cycle using the following procedure: *(i)* Detect the transition point where voltage first exceeds 4.17 V (slightly below the 4.2 V setpoint to account for measurement noise); *(ii)* Track the current profile from this point until current drops below 20 mA or data ends; *(iii)* Validate that the detected CV phase contains at least 20 data points; *(iv)* Verify that current exhibits a decreasing trend during the detected phase. Cycles failing these validation criteria are excluded from analysis. Figure 2 illustrates the CV phase detection process for a representative charging cycle, showing the voltage profile with CC and CV regions in the upper panel and the current profile with exponential decay during the CV phase in the lower panel, where vertical dashed lines mark phase transitions.

To assess the sensitivity of the 4.17 V threshold choice, a sensitivity analysis was conducted across five voltage thresholds (4.15 V, 4.17 V, 4.18 V, 4.19 V, 4.20 V), re-extracting all CV indicators from raw charge cycles for each threshold. The LOBO RMSE varies by only 0.18 percentage points (4.12–4.30%), and $R^2$ remains within 0.793–0.817. All five thresholds detected the same 623 valid cycles, confirming robust CV phase identification. The 4.17 V threshold was retained for its practical balance between early detection and noise robustness. Full results are reported in Table C1.

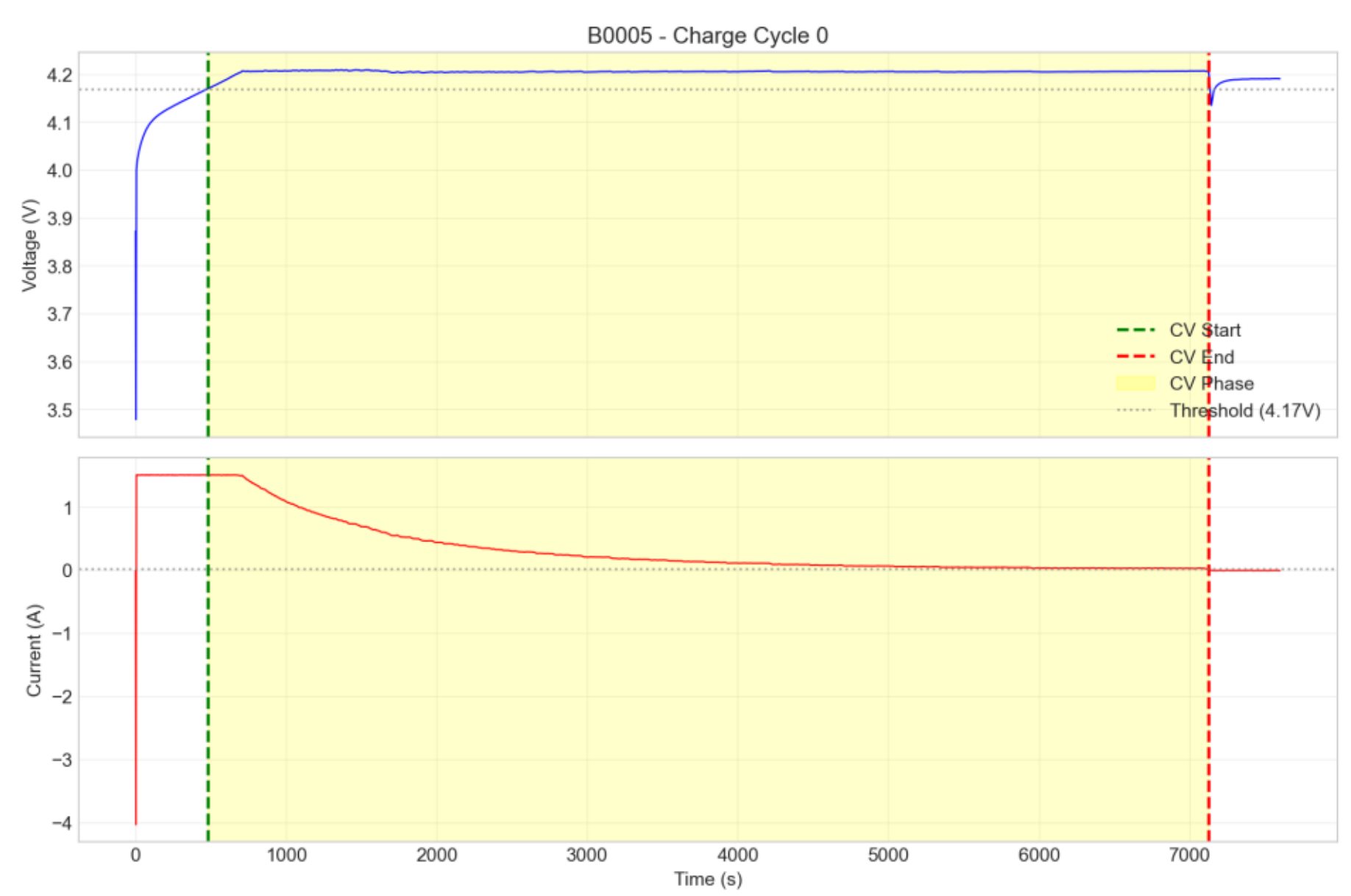

Figure 2. Example of CV phase detection in a charging cycle.

### 3.3. Health indicator extraction

Four health indicators are extracted from the CV charging phase and defined as follows:

**Indicator 1: CV phase duration, denoted $t_{CV}$.**

The time duration of the CV charging phase in seconds,

calculated as $t_{\mathrm{CV}} = t_{\mathrm{end}} - t_{\mathrm{start}}$, where $t_{\mathrm{start}}$ is the time when voltage reaches the CV threshold and $t_{\mathrm{end}}$ is when current drops below the cutoff value.

**Indicator 2: CV-to-CC time ratio ($t_{\mathrm{CV}}/t_{\mathrm{CC}}$).**

The ratio of CV phase duration to CC phase duration. This dimensionless ratio captures the relative time spent in each charging phase, which increases as battery internal resistance grows with aging.

The CC phase duration $t_{\mathrm{CC}}$ is formally defined as the time interval from the start of constant-current charging to the point where voltage reaches the CV threshold (4.17 V): $t_{\mathrm{CC}} = t_{CV,\mathrm{start}} - t_{charge,\mathrm{start}}$. It is extracted as a by-product of the CV phase detection procedure, requiring no additional processing.

**Indicator 3: current decay time constant ($\tau$).**

The CV phase current profile is modeled as exponential decay: $I(t) = I_0 exp(-t/\tau) + I_{\mathrm{offset}}$, where $I_0$ is the initial current amplitude, $\tau$ is the time constant, and $I_{\mathrm{offset}}$ accounts for the cutoff current. The time constant is estimated using nonlinear least squares fitting. When fitting fails to converge, $\tau$ is approximated as the time required for current to decay to 36.8% of its initial value.

In practice, exponential fitting converged for 636 of 637 CV cycles (99.8%). Only 1 cycle (0.2%) in B0018 required the 36.8% decay fallback (per-battery: B0005 0/168, B0006 0/168, B0007 0/168, B0018 1/133). Comparing both methods on the 636 fitted cycles: mean difference (fitted − fallback) = -85.8 s (-6.7%), SD = 59.5 s, max |difference| = 340.7 s. The negative bias indicates the fallback slightly overestimates $\tau$ due to the $I_{\mathrm{offset}}$ term in the exponential model. Given the 99.8% convergence rate, the fallback has negligible dataset-level impact.

**Indicator 4: CV phase charge throughput ($Q_{\mathrm{CV}}$).**

The charge transferred during the CV phase, calculated by integrating current over time. This represents the portion of total charge delivered during the CV phase.

### 3.4. Machine learning model and validation

LightGBM (Light Gradient Boosting Machine) is selected as the prediction model due to its computational efficiency and strong performance on tabular data [27]. The model configuration uses 100 estimators with maximum depth of 6 and learning rate of 0.1. Features are standardized using z-score normalization before model training.

To verify robustness of the default configuration, a grid search was performed over 27 combinations: *n_estimators* ∈ {50, 100, 200}, *max_depth* ∈ {4, 6, 8}, *learning_rate* ∈ {0.05, 0.1, 0.2}, evaluated using 5-fold CV. The best configuration (200, 4, 0.2) achieved RMSE = 1.855%, compared to the default (100, 6, 0.1) at 1.932% — a difference of only 0.08 percentage points (4.0% relative). Given the small dataset (623 samples) and emphasis on practical reproducibility, the default configuration was retained.

Two validation strategies are employed to assess model performance. The primary validation uses LOBO cross-validation, where the model is trained on data from three batteries and tested on the remaining battery, repeated for all four batteries. This approach evaluates cross-battery generalization, which is critical for practical deployment where training and test batteries differ. For comparison with literature results, standard 5-fold cross-validation with random shuffling is also reported. Performance metrics include Root Mean Square Error (RMSE), Mean Absolute Error (MAE), and coefficient of determination ($R^2$).

SHAP (SHapley Additive exPlanations) values are computed to quantify indicator contributions to model predictions [28]. SHAP provides both global importance rankings and local explanations for individual predictions, facilitating interpretation of the model's decision-making process.

## 4. Results

### 4.1. Indicator extraction results

CV phase indicators were successfully extracted from 623 out of 630 charge cycles (98.9% success rate). Seven cycles were excluded due to incomplete data or failure to meet validation criteria.

### 4.2. Indicator correlation and trend analysis

Table 2 presents the Pearson correlation coefficients between extracted indicators and SOH. All four indicators exhibit negative correlation with SOH, indicating that indicator values increase as the battery degrades. The CV phase charge throughput ($Q_{\mathrm{CV}}$) shows the strongest linear correlation ($r$ = -0.758), while the CV-to-CC time ratio and time constant demonstrate equally strong correlations ($r$ = -0.719). The correlations between the extracted CV-phase indicators and SOH, together with their cross-battery consistency, are visualized in Figure 3.

Table 2. Pearson correlation coefficients between CV phase indicators and SOH.

| Indicator | Correlation ($r$) | Interpretation |
|---|---|---|
| $t_{CV}$ | -0.492 | Moderate negative |
| $t_{CV}/t_{CC}$ | -0.719 | Strong negative |
| $\tau$ | -0.719 | Strong negative |
| $Q_{CV}$ | -0.758 | Strong negative |

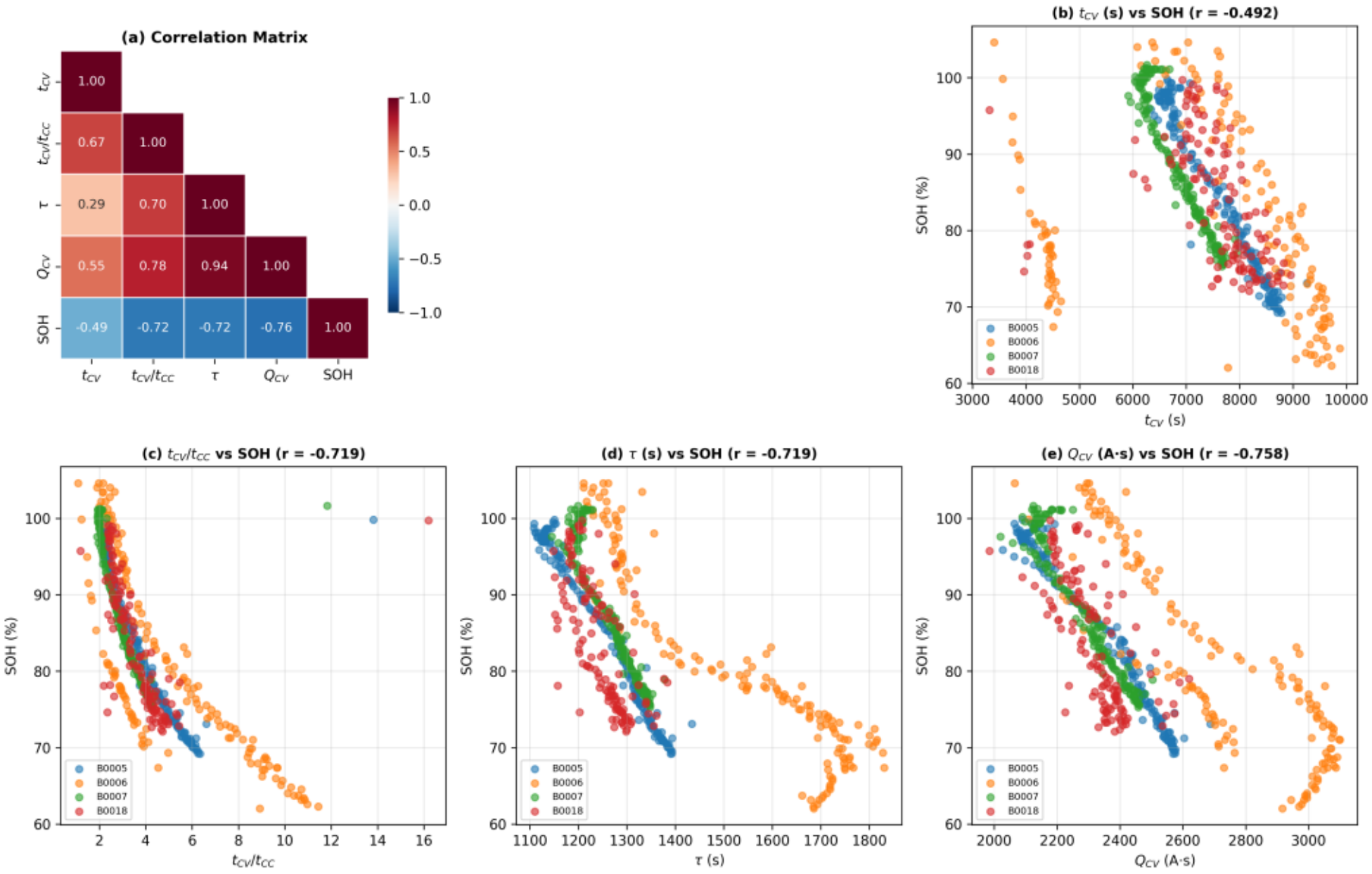

Figure 3. Correlation analysis of CV phase indicators: (a) Correlation heatmap showing relationships between all indicators and SOH; (b–e) Scatter plots of key indicators versus SOH, colored by battery ID, demonstrating consistent degradation trends across different batteries.

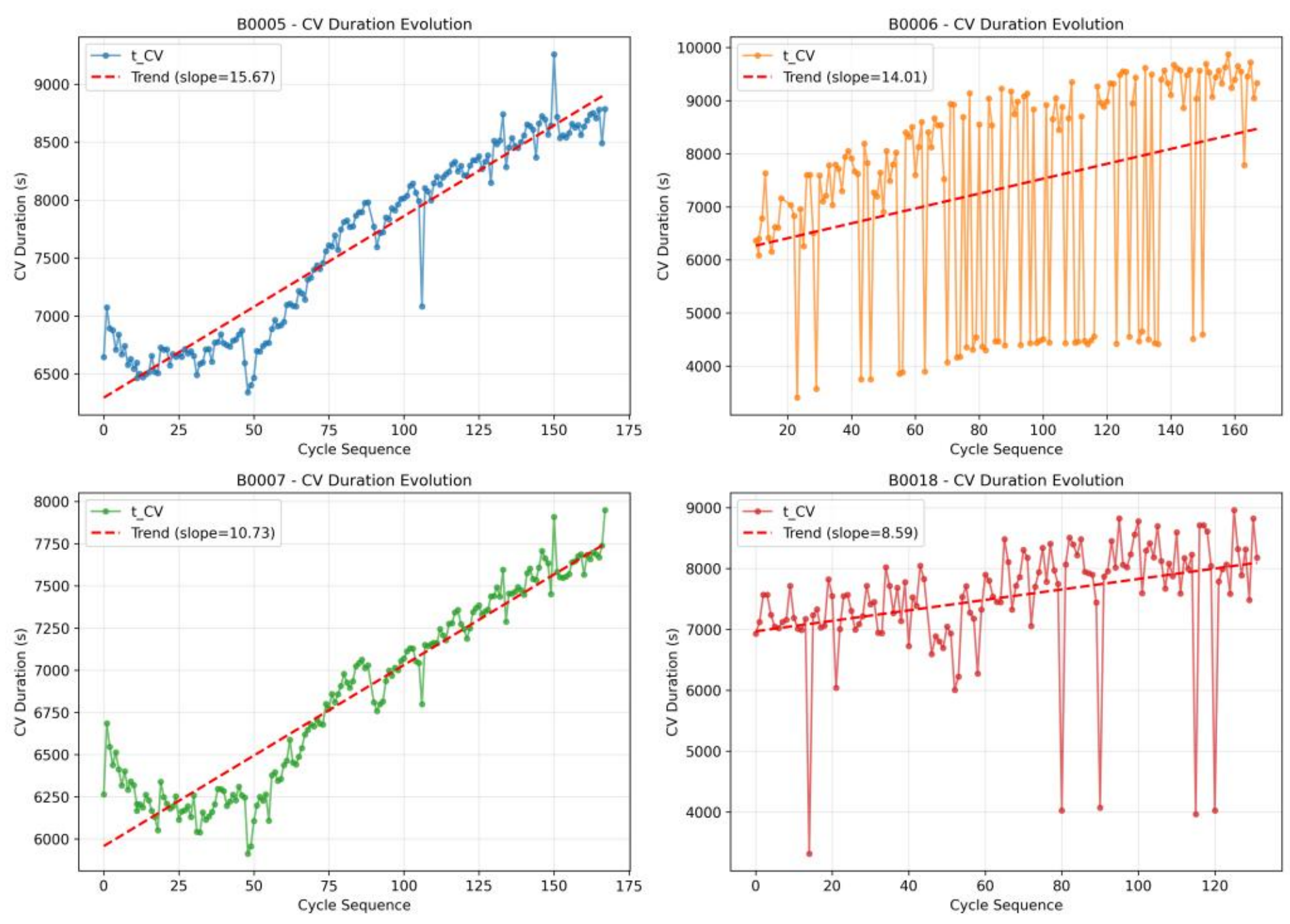

Figure 4. Evolution of CV phase duration over battery cycling

The evolution of CV phase duration over battery cycling is analyzed to assess monotonicity. Figure 4 shows that all four batteries exhibit increasing CV duration trend with cycling, consistent with the expected physical behavior of increasing internal resistance during aging. Table 3 summarizes the monotonicity analysis results, confirming positive slopes ranging from 8.69 to 15.76 seconds per cycle. The detailed statistical distribution of all indicators across batteries is provided in Appendix A.

Table 3. CV duration monotonicity analysis showing trend slopes and the percentage of increasing steps.

| Battery | Slope (s/cycle) | Percentage of increasing steps (%) |
|---|---|---|
| B0005 | 15.76 | 61.1% |
| B0006 | 14.23 | 52.3% |
| B0007 | 10.79 | 56.3% |
| B0018 | 8.69 | 47.7% |

Table 4. Five-fold cross-validation results.

| Model | RMSE (%) | MAE (%) | $R^2$ |
|---|---|---|---|
| Random Forest | 2.12 ± 0.48 | 1.18 ± 0.10 | 0.958 ± 0.020 |
| XGBoost | 2.22 ± 0.51 | 1.21 ± 0.11 | 0.954 ± 0.022 |
| CatBoost | 2.23 ± 0.50 | 1.20 ± 0.10 | 0.954 ± 0.021 |
| LightGBM | 2.30 ± 0.52 | 1.25 ± 0.12 | 0.951 ± 0.023 |

### 4.3. Model performance

#### 4.3.1. Five-fold cross-validation results

Table 4 presents the 5-fold cross-validation results comparing multiple baseline models. To assess the generality of results across different learning algorithms, four models were evaluated: LightGBM, XGBoost, CatBoost, and Random Forest. Under standard cross-validation, all models achieve comparable performance (RMSE: 2.12–2.30%, $R^2$: 0.951–0.958), with Random Forest showing marginally the lowest RMSE.

#### 4.3.2. Leave-One-Battery-Out results

Table 5 presents the more rigorous LOBO validation results. The LOBO results reveal substantial performance variation across batteries. B0005 and B0007 achieve $R^2$ values above 0.83, while B0006 and B0018 show lower accuracy. The overall RMSE of 4.69% and $R^2$ of 0.796 represent realistic expectations for cross-battery deployment. Figure 5 presents the SOH prediction results under LOBO validation for all four batteries, highlighting battery-specific performance variations.

Across the four LOBO folds, the variability is: RMSE = 4.69 ± 1.26%, MAE = 3.69 ± 1.03%, $R^2$ = 0.769 ± 0.110. The relatively large standard deviations reflect heterogeneous degradation behaviors: B0005 and B0007 achieve RMSE ≈ 3.3–3.5%, while B0006 and B0018 exhibit RMSE ≈ 5.2–6.4%. These variability estimates should be interpreted with caution given the small number of folds (N = 4).

Table 5. LOBO validation results, representing realistic cross-battery generalization performance.

| Test battery | Number of samples | RMSE (%) | MAE (%) | $R^2$ |
|---|---|---|---|---|
| B0005 | 168 | 3.30 | 2.62 | 0.896 |
| B0006 | 156 | 6.37 | 5.19 | 0.732 |
| B0007 | 168 | 3.48 | 2.93 | 0.839 |
| B0018 | 131 | 5.18 | 4.24 | 0.609 |
| Overall | 623 | 4.69 ± 1.26% | 3.69 ± 1.03% | 0.769 ± 0.110 |

#### 4.3.3. Model comparison

To assess whether the observed cross-battery performance gap stems from model limitations or inherent generalization challenges, multiple baseline models were evaluated under identical LOBO validation. Table 6 presents the results. All four gradient boosting methods achieve comparable LOBO performance (RMSE: 4.63–5.08% and $R^2$: 0.760–0.801), with LightGBM showing marginally the best accuracy. The narrow performance range across fundamentally different algorithms suggests that cross-battery generalization, rather than model capacity, represents the primary bottleneck. This finding has practical implications: practitioners can select models based on deployment requirements without sacrificing significant accuracy.

Table 6. Model comparison under LOBO validation, showing comparable performance across gradient boosting methods and indicating that cross-battery generalization is the primary bottleneck.

| Model | RMSE (%) | MAE (%) | $R^2$ |
|---|---|---|---|
| LightGBM | 4.63 | 3.60 | 0.801 |
| XGBoost | 4.65 | 3.64 | 0.799 |
| CatBoost | 5.07 | 4.10 | 0.761 |
| Random Forest | 5.08 | 4.08 | 0.760 |

### 4.4. Indicator set comparison

In Table 7, four indicator sets are compared. "CV-only" refers to the four indicators extracted solely from the CV phase: CV duration, CV/CC time ratio, current decay time constant, and CV-phase charge throughput. "CV-ratio only" uses only the single CV/CC time ratio indicator. "$t_{CC}$ only" denotes the total duration of the constant-current phase as a single indicator.

"Combined (CV+ $t_{CC}$)" includes all four CV-based indicators together with the CC-phase duration. The column "N indicators" indicates the number of indicators used in each set. The combination of CV and CC indicators yields the best performance ($R^2$ = 0.874). Notably, the single CC phase duration indicator ($t_{CC}$) achieves slightly higher accuracy than the four CV indicators combined ($R^2$ of 0.845 versus 0.796). However, the combined approach (CV + CC) achieves the best overall performance ($R^2$ = 0.874), indicating that CV indicators capture complementary degradation information not fully represented by CC duration alone.

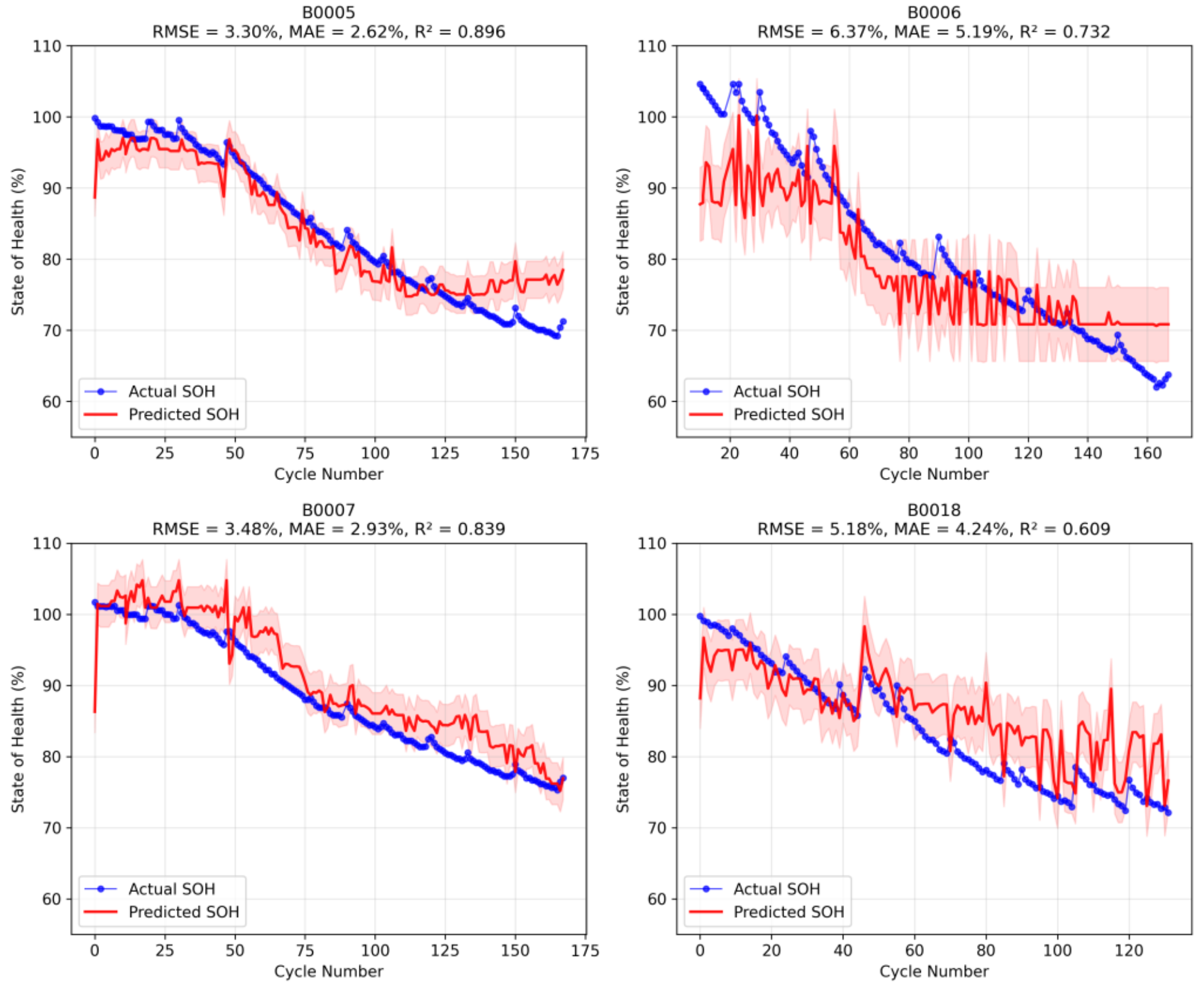

Figure 5. SOH prediction results under LOBO validation for all four batteries.

Table 7. Comparison of different indicator sets under LOBO validation.

| Indicator set | N indicators | RMSE (%) | MAE (%) | $R^2$ |
|---|---|---|---|---|
| CV-only | 4 | 4.69 | 3.69 | 0.796 |
| CV-ratio only | 1 | 5.55 | 4.14 | 0.715 |
| $t_{CC}$ only | 1 | 4.09 | 2.72 | 0.845 |
| Combined (CV + $t_{CC}$) | 5 | 3.68 | 2.87 | 0.874 |

### 4.5. Indicator importance analysis

SHAP analysis identifies the CV-to-CC time ratio as the most influential indicator for SOH prediction (Figure 6). The ratio's top ranking is notable because it normalizes for variations in charging conditions and provides a dimensionless indicator that may generalize better across different battery systems. Table 8 summarizes the SHAP-based importance ranking. The overall agreement between predicted and measured SOH values under LOBO validation is shown in Figure 7. The dashed diagonal line represents perfect prediction. The gray band indicates ±2% error region. The prediction error distribution analysis is provided in Appendix B.

Table 8 is updated with numerical mean |SHAP| values: $t_{CV}/t_{CC}$ = 7.839 (Highest), $t_{CV}$ = 1.626 (High), $Q_{CV}$ = 1.346 (Moderate), $\tau$ = 1.275 (Lower). The $t_{CV}/t_{CC}$ ratio dominates at approximately 4.8× the next indicator. Under LOBO validation: $t_{CV}/t_{CC}$= 6.77 ± 1.56, $Q_{CV}$ = 2.45 ± 2.15, $\tau$ = 2.32 ± 2.05, $t_{CV}$ = 1.38 ± 0.67.

Table 8. SHAP-based indicator importance ranking.

| Rank | Indicator | Mean \|SHAP\| | Relative Importance |
|---|---|---|---|
| 1 | $t_{CV}/t_{CC}$ | 7.839 | Highest |
| 2 | $t_{CV}$ | 1.626 | High |
| 3 | $Q_{CV}$ | 1.346 | Moderate |
| 4 | $\tau$ | 1.275 | Lower |

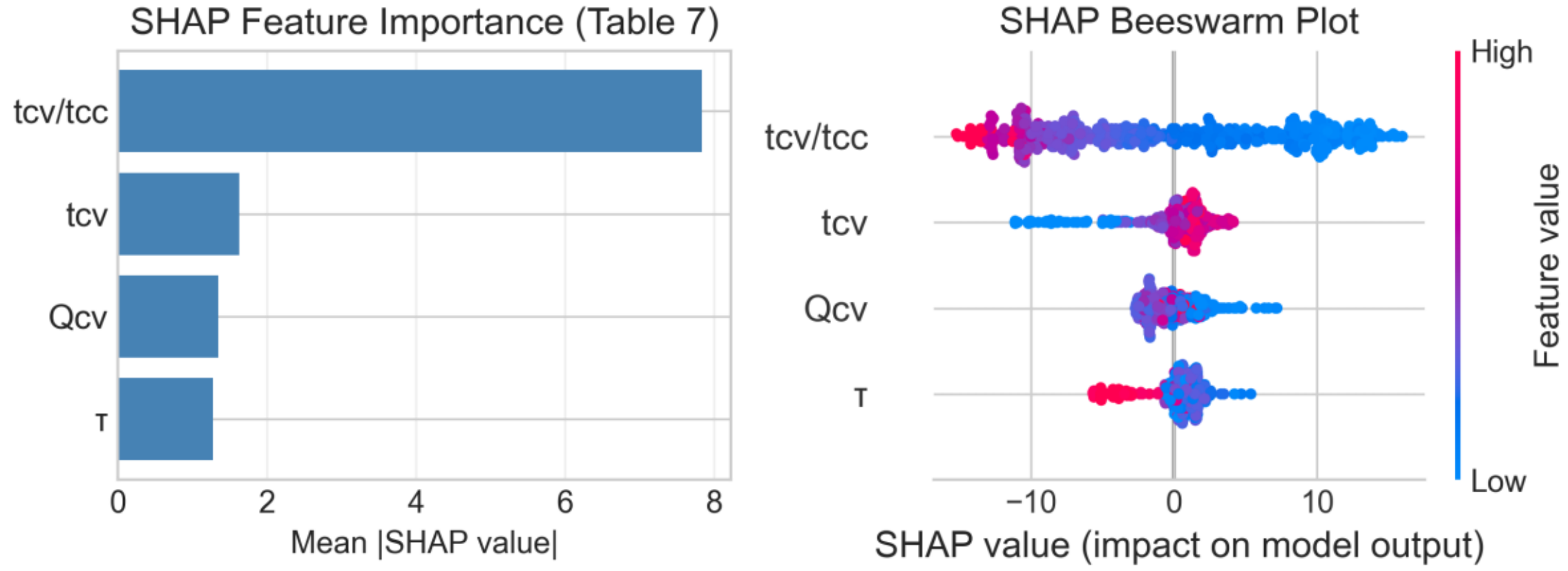

Figure 6. SHAP indicator importance analysis showing mean absolute SHAP value.

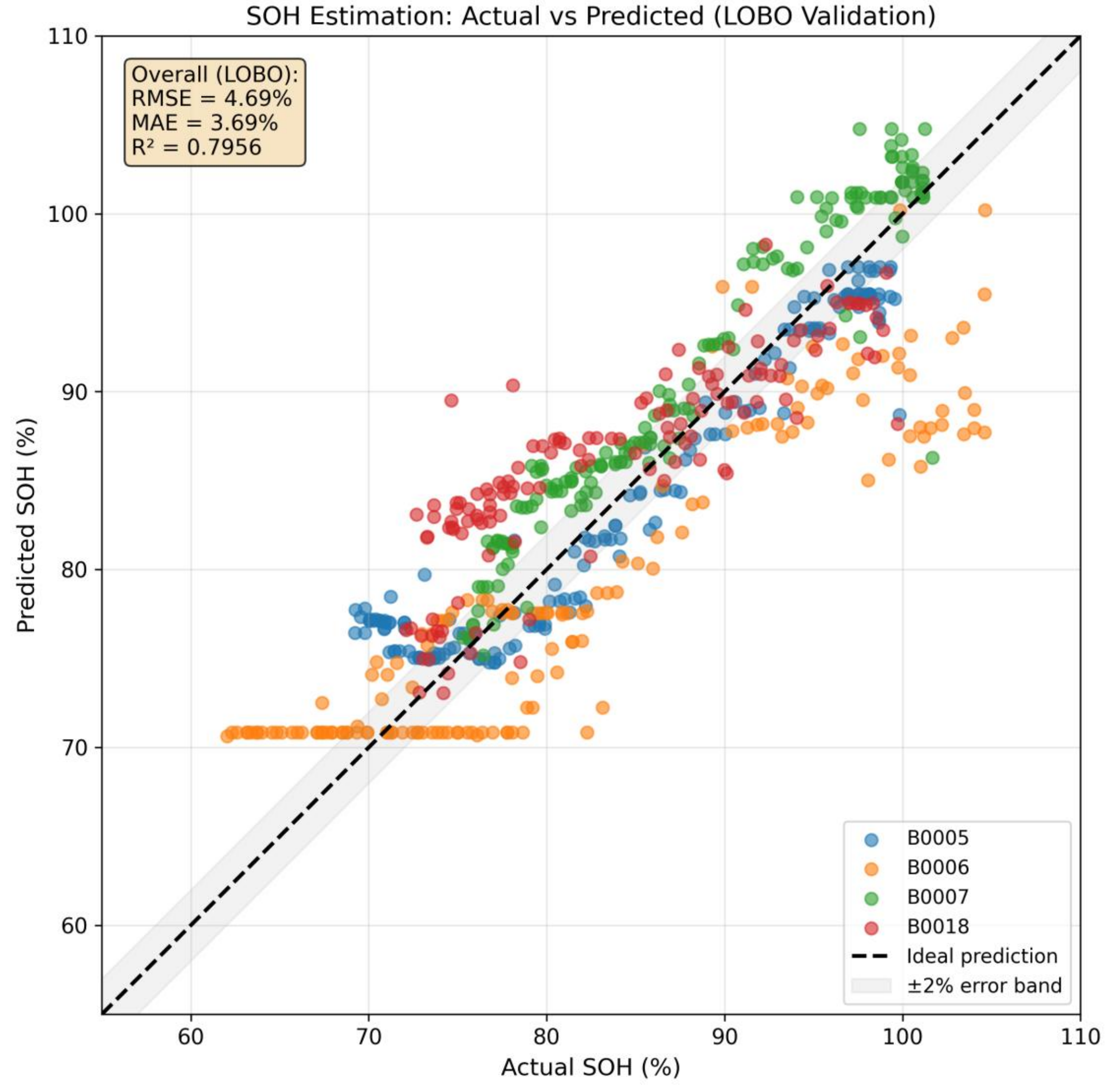

Figure 7. Actual versus predicted SOH for all batteries under LOBO validation.

## 5. Discussion

### 5.1. Practical implications for maintenance

The achieved RMSE of 4.69% under LOBO validation may appear modest compared to state-of-the-art deep learning methods reporting sub-1% errors. However, from a maintenance engineering perspective, this accuracy level remains practically useful for several reasons. Recent studies on data-driven predictive maintenance strategies have emphasized that accurate failure prognostics enable optimal timing of maintenance activities [29], while machine learning-based decision support systems can effectively guide maintenance function improvement even with moderate estimation accuracy [30].

First, regarding maintenance threshold decisions, battery replacement is typically triggered when SOH drops below 70–80%. A 5% estimation error is unlikely to cause premature replacement or delayed maintenance around this threshold. Second, the monotonic relationship between CV indicators and SOH enables reliable degradation trend tracking, even if absolute accuracy is limited. Maintenance planners can use trend information for remaining useful life prediction. Third, for applications where computational resources are constrained, accepting moderate accuracy in exchange for simpler indicator extraction and model implementation may be a rational engineering trade-off.

### 5.2. Cross-battery generalization challenge

The significant performance gap between 5-fold CV and LOBO validation highlights the challenge of cross-battery generalization. Table 9 quantifies this gap across all evaluated models. On average, LOBO validation yields RMSE approximately 119% higher than standard 5-fold CV, demonstrating that random train-test splits substantially overestimate practical deployment performance. This gap likely stems from battery-to-battery variation due to manufacturing tolerances and subtle differences in aging conditions, feature distribution shift where the statistical distribution of CV indicators may differ across batteries, and limited training diversity with only three batteries available for training in LOBO.

Table 9. Comparison of validation strategies, showing substantially higher RMSE for LOBO validation than for 5-fold cross-validation.

| Model | 5-fold RMSE (%) | LOBO RMSE (%) | RMSE increase |
|---|---|---|---|
| LightGBM | 2.30 | 4.63 | +101% |
| XGBoost | 2.22 | 4.65 | +110% |
| CatBoost | 2.23 | 5.07 | +127% |
| Random Forest | 2.12 | 5.08 | +140% |
| **Average** | **2.22** | **4.86** | **+119%** |

As shown in Appendix A, B0006 exhibits notably higher variability in indicator values (e.g., $t_{\mathrm{CV}}$ standard deviation of 2014 s compared to 554–945 s for other batteries), which may explain its lower prediction accuracy under LOBO validation.

Detailed per-battery error analysis (Table B2) reveals: B0005 (mean error -0.16%, SD 3.30%), B0006 (-2.55%, SD 5.83%), B0007 (+2.61%, SD 2.30%), B0018 (+2.56%, SD 4.50%). B0006 exhibits the highest $t_{\mathrm{CV}}$ variability (SD = 2014 s vs 554–945 s for others), $\tau$ variability (SD = 203 s), and $Q_{CV}$ variability (SD = 286 A·s), making cross-battery generalization inherently difficult. B0018 has the fewest cycles (131 vs 156–168), SOH up to 104.6%, and lowest $t_{\mathrm{CV}}$ monotonicity (47.7% increasing steps), explaining its elevated prediction errors.

These findings underscore the importance of validating battery SOH methods using cross-battery or cross-condition protocols rather than random train-test splits that may overestimate practical performance.

### 5.3. Comparison of CV and CC indicators

The observation that CC phase duration alone ($t_{\mathrm{CC}}$) slightly outperforms the four CV indicators combined ($R^2$ of 0.845 vs 0.796) deserves careful interpretation. CC phase duration directly reflects capacity-related aging, as the time to charge to a fixed voltage threshold depends on available capacity. CV phase indicators may be more sensitive to internal resistance changes, which do not perfectly correlate with capacity fade.

*The finding that the combined approach ($R^2$ = 0.874) substantially outperforms both CC-only ($R^2$ = 0.845) and CV-only ($R^2$ = 0.796) approaches demonstrates that the two charging phases capture complementary degradation information.* CC phase duration primarily reflects capacity-related aging through the time required to reach the voltage threshold, while CV phase indicators are more sensitive to internal resistance changes manifested in current decay characteristics. The CV-to-CC time ratio, identified by SHAP analysis as the most influential indicator, serves as a dimensionless measure that normalizes for variations in charging conditions and captures the shift from CC-dominated to CV-dominated charging as batteries age. This complementarity explains why combining indicators from both phases yields superior prediction performance. From a practical deployment perspective, the choice of indicator set depends on data availability and system constraints: *(i)* when complete CC-CV charging data is available, the combined approach is recommended for optimal accuracy; *(ii)* when only partial

charging data is logged or CC phase is disrupted by adaptive charging protocols, CV-phase indicators provide a robust alternative due to their smooth profiles requiring no numerical differentiation; *(iii)* for embedded BMS implementations with limited computational resources, CV indicators offer advantages in terms of extraction simplicity and physical interpretability for maintenance personnel.

### 5.4. Physical interpretation of CV indicators

The physical significance of the evaluated indicators enhances their credibility for maintenance applications. CV duration increase reflects impedance rise, as more time is required to complete charging in the CV phase when internal resistance grows. CV/CC ratio increase indicates reduced charge acceptance capability, showing the shift from CC-dominated to CV-dominated charging. Time constant increase reflects increased diffusion resistance in the electrode materials during aging. CV charge throughput increase shows that more charge is delivered during CV phase to compensate for reduced CC phase capacity. These physically interpretable relationships provide confidence that the indicators capture genuine aging phenomena rather than spurious correlations.

### 5.5. Limitations

Several limitations should be acknowledged. First, the NASA dataset represents controlled laboratory conditions with a single cell chemistry ($LiCoO_2$). Real-world applications involve diverse chemistries, form factors, and operating conditions that may affect indicator behavior. Second, the evaluated indicators depend on the CV phase cutoff current setting. Different BMS configurations may use different thresholds, requiring indicator recalibration. This operational dependency should be considered when deploying the method across different systems. Third, the dataset was collected at approximately 24 °C. CV phase behavior varies with temperature, potentially affecting indicator reliability in variable-temperature environments. Fourth, preliminary analysis indicates that models trained on early cycles cannot reliably predict late-cycle SOH (negative $R^2$ in chronological split validation). *This limitation is consistent with recent findings that battery aging trajectories are path-dependent and difficult to extrapolate using early-life data alone, even with more complex deep learning models [17].* The proposed indicators are therefore most suitable for relative health monitoring and trend tracking within individual battery systems, rather than absolute SOH prediction for unseen degradation states. Fifth, results are specific to the CC-CV charging protocol used. Different charging strategies may exhibit different CV phase characteristics.

Sixth, all experiments are conducted on a single dataset (NASA). Cross-dataset validation using datasets such as CALCE, Oxford Battery Degradation, and Stanford/MIT fast-charging would substantially strengthen generalizability claims. However, different CC-CV parameters, sampling rates, and cell chemistries across datasets require indicator recalibration and threshold adjustment. Cross-dataset validation is identified as a priority for future work.

### 5.6. Recommendations for implementation

Based on the findings, the following recommendations are offered for practical implementation. When deploying to new batteries, expect RMSE of 4–5% rather than the optimistic 2% from standard cross-validation. Rather than relying solely on point estimates, track indicator trends over time to improve maintenance decision confidence. If possible, collect initial cycling data from each battery to calibrate the model, potentially improving accuracy for that specific unit. Include CC phase duration when available, as the combination of CV and CC indicators provides better performance than either alone.

From a deployment perspective, Table 10 summarizes the computational characteristics of the evaluated models. All gradient boosting methods achieve inference times below 2 milliseconds per sample, enabling real-time SOH estimation at typical BMS sampling rates (1–10 Hz). Model sizes remain below 250 kB, compatible with resource-constrained microcontrollers commonly used in battery management systems. LightGBM offers the best balance between accuracy (lowest LOBO RMSE) and computational efficiency, making it the recommended choice for practical embedded implementation.

Table 10. Computational characteristics of the models for embedded deployment, including inference time and model size.

| Model | Inference time (ms) | Model size (kB) |
|---|---|---|
| LightGBM | 1.4 | 111 |
| XGBoost | 0.6 | 229 |
| CatBoost | 1.4 | 65 |
| Random Forest | 32.9 | 3106 |

Table 10 now includes Random Forest for completeness:

inference time 32.9 ms, model size 3106 KB. Random Forest is 23–55× slower in inference and 13–48× larger in model size compared to the gradient boosting models, while achieving comparable LOBO accuracy (RMSE = 5.08%). This further supports LightGBM as the recommended model for resource-constrained BMS deployments.

To facilitate practical deployment, Table 11 provides an indicator selection guideline based on data availability and system constraints. This decision framework allows maintenance engineers to select the most appropriate indicator set for their specific operational context.

Table 11. Indicator selection guideline based on data availability and operational constraints.

| Data condition | Recommended indicators | Rationale |
|---|---|---|
| Complete CC-CV data available | Combined (CV + $t_{CC}$) | Highest accuracy ($R^2$ = 0.874) |
| Only CV phase logged | CV duration, $\tau$, $Q_{CV}$ | Smooth extraction, no differentiation |
| CC phase unstable/adaptive | CV-only indicators | CC duration loses meaning with variable protocols |
| Embedded/low-power BMS | CV duration or CV/CC ratio | Minimal computation, noise-robust |
| Degradation trend monitoring | CV duration | Strong monotonicity (8.7–15.8 s/cycle increase) |

As a practical maintenance example, consider a battery system where SOH degrades from 100% to the 80% replacement threshold. Based on Table 3, CV phase duration increases at approximately 8.7–15.8 seconds per cycle. For a battery completing 168 cycles (similar to B0005 in this study), the CV duration would increase from approximately 2,800 seconds to over 4,500 seconds. A maintenance engineer could therefore set a threshold-based alarm: when CV duration exceeds 4,000 seconds, schedule battery inspection; when it exceeds 4,500 seconds, plan replacement. This approach requires only simple time measurements without complex SOH estimation algorithms, making it suitable for field deployment where computational resources are limited.

Unlike differential features such as $dQ/dV$ or $dI/dt$ that require numerical differentiation and are sensitive to sampling rate and measurement noise, CV-phase indicators can be extracted using only time and current logs with simple threshold detection. This extraction simplicity, combined with the physically interpretable relationship between CV duration and internal resistance growth, makes CV indicators particularly attractive for maintenance personnel who need actionable health information without deep machine learning expertise.

## 6. Conclusions

This study conducted a systematic comparison of CC-only, CV-only, and combined charging phase indicators for battery SOH estimation using rigorous LOBO validation. Four CV-phase indicators were investigated, namely CV phase duration, CV-to-CC time ratio, current decay time constant, and CV phase charge throughput.

Among these indicators, the CV-to-CC time ratio was identified as the most influential feature through SHAP analysis. As a dimensionless indicator, it offers clear physical interpretability by reflecting the increase in internal resistance and reduced charge acceptance capability associated with battery aging. All evaluated CV-phase indicators exhibited monotonic relationships with degradation, with CV phase duration increasing at rates of 8.7–15.8 seconds per cycle across the evaluated batteries, consistent with expected aging behavior.

Using the rigorous LOBO validation protocol, the CV-only approach achieved an RMSE of 4.69% and an $R^2$ of 0.796, providing a realistic assessment of cross-battery generalization performance. The combination of CV and CC phase indicators further improved prediction accuracy ($R^2$ = 0.874), indicating that the two charging phases capture complementary degradation information. In contrast, the substantial performance gap between standard cross-validation and LOBO validation highlights the extent to which commonly used random splits can overestimate deployable SOH estimation performance, underscoring the need for more stringent evaluation protocols in practical battery health monitoring studies.

From a reliability engineering perspective, this study provides practical deployment guidance through an indicator selection framework (Table 11) that helps maintenance engineers choose appropriate indicators based on data availability and system constraints. The key engineering insights are: *(i)* use combined CC+CV indicators for maximum accuracy when complete charging data is available; *(ii)* rely on CV-only indicators when CC phase is unstable or when computational simplicity is required; *(iii)* implement threshold-based alarms on CV duration for simple field deployment without complex algorithms. Unlike differential features requiring numerical differentiation, CV indicators can be extracted from basic time and current measurements, making them particularly suitable for embedded battery management systems where computational resources are limited and robustness to measurement noise is essential.

Future work will focus on validating these charging phase indicators across diverse battery chemistries and real-world operating conditions, developing adaptive strategies to accommodate varying CV termination settings, and integrating the indicators with remaining useful life prediction frameworks to support comprehensive maintenance planning in operational energy storage systems.

## Appendix A: Detailed indicator statistics

Table A1 presents the detailed statistical distribution of CV phase indicators across all four batteries. These statistics provide additional context for understanding the indicator behavior and variability observed in the main analysis.

**Table A1.** Detailed indicator statistics by battery (mean ± standard deviation).

| Battery | $t_{CV}$ (s) | $t_{CV}/t_{CC}$ | $\tau$ (s) | $Q_{CV}$ (A·s) |
|---|---|---|---|---|
| B0005 | 7595 ± 800 | 3.821 ± 1.588 | 1254 ± 95 | 2338 ± 183 |
| B0006 | 7373 ± 2014 | 4.946 ± 2.791 | 1520 ± 203 | 2706 ± 286 |
| B0007 | 6849 ± 554 | 2.933 ± 1.032 | 1259 ± 56 | 2278 ± 118 |
| B0018 | 7532 ± 945 | 3.623 ± 1.433 | 1244 ± 52 | 2320 ± 104 |

## Appendix B: Prediction error analysis

Figure B1, Table B1, and Table B2 present the detailed prediction error analysis under LOBO validation. The error distribution is approximately symmetric with mean error of -0.12% (slight underestimation bias) and standard deviation of 4.68%. The 95th percentile absolute error is 9.8%, with outliers primarily from the B0006 and B0018 batteries.

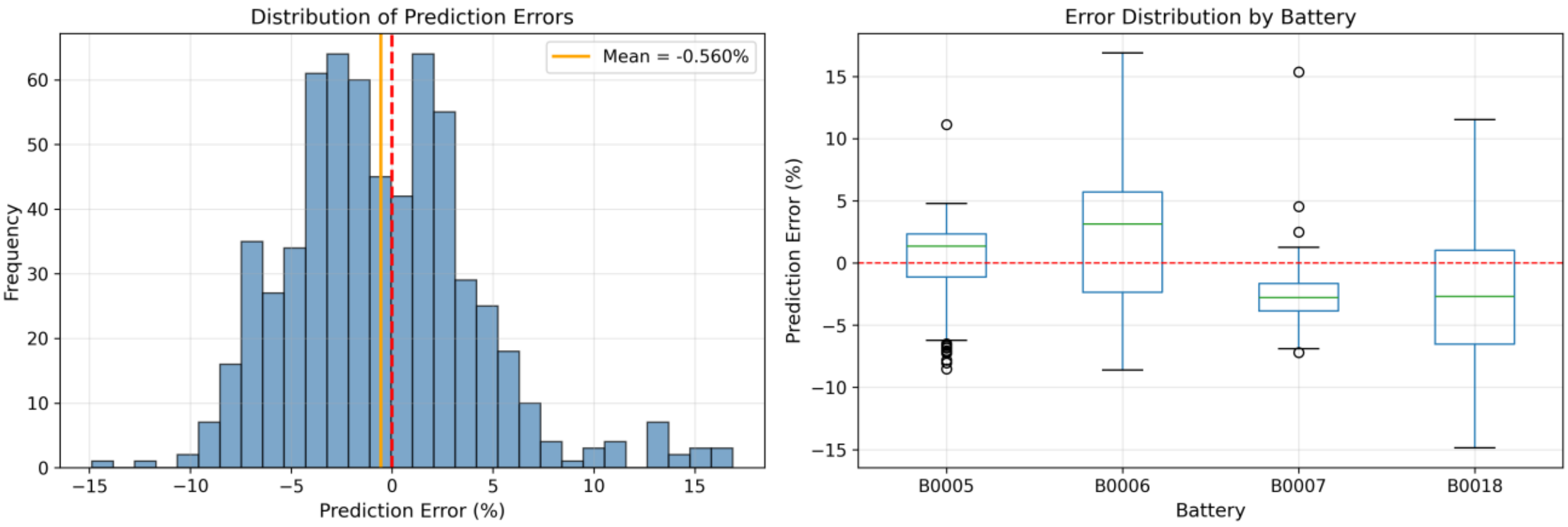

**Figure B1.** Prediction error analysis: (a) Histogram of prediction errors showing approximately symmetric distribution around zero; (b) Box plots of prediction errors by battery, showing variation in prediction accuracy across batteries. B0006 and B0018 exhibit larger error variance compared to B0005 and B0007.

**Table B1.** Prediction error statistics under LOBO validation.

| Statistic | Value |
|---|---|
| Mean error | -0.12% |
| Standard deviation | 4.68% |
| Maximum absolute error | 15.2% |
| 95th percentile \|error\| | 9.8% |
| Median absolute error | 3.1% |

**Table B2.** Per-battery prediction error statistics under LOBO validation (LightGBM).

| Battery | Mean Err (%) | Std (%) | Max \|Err\| (%) | P95 (%) | RMSE (%) |
|---|---|---|---|---|---|
| B0005 | -0.16 | 3.30 | 11.12 | 6.98 | 3.30 |
| B0006 | -2.55 | 5.83 | 16.88 | 13.57 | 6.37 |
| B0007 | +2.61 | 2.30 | 15.37 | 5.98 | 3.48 |
| B0018 | +2.56 | 4.50 | 14.87 | 8.74 | 5.18 |

**Appendix C: Threshold sensitivity analysis**

**Table C1.** Sensitivity analysis across five voltage thresholds

| Threshold (V) | Cycles | LOBO RMSE (%) | LOBO $R^2$ |
|---|---|---|---|
| 4.15 | 623 | 4.300 | 0.793 |
| 4.17 (baseline) | 623 | 4.235 | 0.808 |
| 4.18 | 623 | 4.121 | 0.817 |
| 4.19 | 623 | 4.237 | 0.812 |
| 4.20 | 623 | 4.221 | 0.814 |